\documentclass[letterpaper, 10 pt, journal, twoside]{IEEEtran/IEEEtran}
\usepackage{packages}

\author{Manuel Castillo-Lopez$^{1}$, Philippe Ludivig$^{1,2}$, Seyed Amin Sajadi-Alamdari$^{1}$,\\ Jose Luis Sanchez-Lopez$^{1}$, Miguel A. Olivares-Mendez$^{1}$, Holger Voos$^{1}$
\thanks{Manuscript received: September 10, 2019; Revised: November 13, 2019; Accepted: January 19, 2020.}
\thanks{This paper was recommended for publication by Editor Nancy Amato upon evaluation of the Associate Editor and Reviewers' comments.
This work was supported by the ''Fonds National de la Recherche'' (FNR), Luxembourg, under the projects C15/15/10484117 (BEST-RPAS).)} 
\thanks{$^{1}$The authors are with the Automation and Robotics Research Group, Interdisciplinary Centre for Security, Reliability and Trust (SnT), University of Luxembourg, Luxembourg
  {\tt\footnotesize \{\href{mailto:manuel.castillo@uni.lu}{manuel.castillo}, \href{mailto:holger.voos@uni.lu}{holger.voos}\}@uni.lu}}%
\thanks{$^{2}$The Second author is also with ispace Europe, Luxembourg \href{mailto:p-ludivig@ispace-inc.com}{\tt\footnotesize p-ludivig@ispace-inc.com}}
\thanks{Digital Object Identifier (DOI): 10.1109/LRA.2020.2975759.}
}
\title{
  A Real-Time Approach for Chance-Constrained Motion Planning with Dynamic Obstacles.
}

\begin{document}

\maketitle

\begin{abstract}

  Uncertain dynamic obstacles, such as pedestrians or vehicles, pose a major challenge for optimal robot navigation with safety guarantees. Previous work on optimal motion planning has employed two main strategies to define a safe bound on an obstacle's space: using a polyhedron or a nonlinear differentiable surface. The former approach relies on disjunctive programming, which has a relatively high computational cost that grows exponentially with the number of obstacles. The latter approach needs to be linearized locally to find a tractable evaluation of the chance constraints, which dramatically reduces the remaining free space and leads to over-conservative trajectories or even unfeasibility. In this work, we present a hybrid approach that eludes the pitfalls of both strategies while maintaining the original safety guarantees. The key idea consists in obtaining a safe differentiable approximation for the disjunctive chance constraints bounding the obstacles. The resulting nonlinear optimization problem can be efficiently solved to meet fast real-time requirements with multiple obstacles. We validate our approach through mathematical proof, simulation and real experiments with an aerial robot using nonlinear model predictive control to avoid pedestrians.

\end{abstract}

\vspace{-0.5cm}
\section{Introduction}\label{sec:introduction}
\IEEEPARstart{A}{utonomous}
robots, such as self-driving cars or drones, are expected to revolutionize transportation, inspection and many other applications to come \cite{mohamed2018unmanned}. To fully exploit their capabilities, we need to enable their safe operation among humans and other robots while pursuing high-level objectives such as safety \cite{hentzen2018maximizing} or energy consumption \cite{sajadi2019nonlinear}. However, planning trajectories with obstacles whose present and future location is highly uncertain is still a difficult and computationally expensive problem \cite{blackmore2011chance}. Strong assumptions need to be made to find tractable solutions for fast real-time applications. This leads to over-conservative obstacle models that ensure collision-free operation but drastically reduce the remaining free space \cite{kamel2017robust,zhu2019chance}, compromising the problem's feasiblity when multiple obstacles arise. As a result, reducing conservatism in motion planning algorithms while providing safety guarantees has become a major problem and the subject of active research \cite{blackmore2011chance, ono2012closed, ono2013probabilistic, jha2018safe, zhu2019chance,lefkopoulos2019using,ono2008efficient}.

In this paper, we present a new approach to model uncertain dynamic obstacles for fast real-time motion planning applications. This method eludes the over-conservatism of existing real-time approaches while providing safety guarantees at a low computational cost. The resulting problem is modeled within the framework of disjunctive chance-constrained optimization and casted into non-linear programming, for which efficient solvers exist \cite{houska2011}. Thus, the main contributions of this paper are listed as follows:
\begin{figure}[t]
    \centering
    \includegraphics[width=\columnwidth]{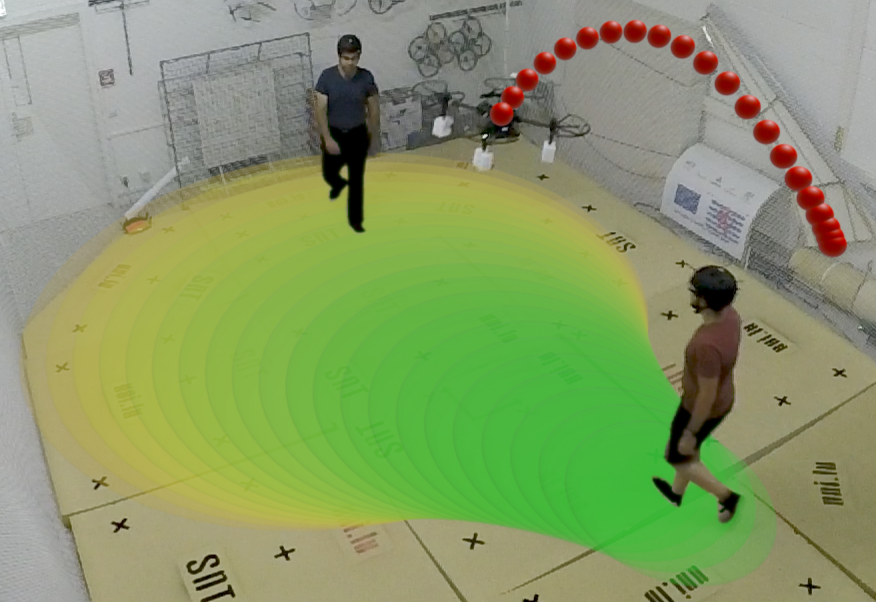}
    \caption{ An instant of the proposed approach running on an aerial vehicle to avoid pedestrians in a cluttered environment. The planned trajectory has been rendered into the image plane as red balls. The predicted bounding ellipsoids of one pedestrian are projected into the ground as degraded-green ellipses. Video: \url{http://rebrand.ly/castillo_RAL2020}}
    \label{fig:experiment_snapshot}
 \vspace{-0.5cm}
\end{figure}

\begin{itemize}
  \item Theoretical results on disjunctive chance constraints, providing tighter bounds on the probability of collision.
  \item A new real-time approach for chance-constrained motion planning in dynamic environments.
  \item Empirical validation through simulation and real experiments on an aerial robot to avoid pedestrians.
\end{itemize}
The rest of the paper is organized as follows: \cref{sec:related_work} presents an overview of existing approaches for chance-constrained motion planning. We formalize the motion planning problem in \cref{sec:problem_statement}, and present a set of preliminary results in \cref{sec:existing} from which we build our theoretical results in \cref{sec:diff_bound}. Finally, our approach is evaluated through a benchmark, software-in-the-loop simulations and real experiments in \cref{sec:case_study}, drawing the resulting conclusions and future lines of work in \cref{sec:conclusions}. 

\section{Related Work}
\label{sec:related_work}

Optimal motion planning has been the subject of active research during the last decade, as surveyed in  \cite{dadkhah2012survey, hoy2015algorithms}. Generally, the space occupied by obstacles is represented as a set of constraints on the free space which, in general, disrupts its convexity. The choice on the type of constraints determines the nature of the resulting optimization problem and therefore, its performance. There are two main strategies in the literature to encapsulate an obstacle's space: Using a convex polyhedron \cite{blackmore2011chance, lefkopoulos2019using} (e.g. a cuboid), or a single differentiable surface \cite{castillo2018model, kamel2017robust, zhu2019chance} (e.g. an ellipsoid).

A polyhedral obstacle is encoded as a disjunction of linear inequality constraints. This represents logical \textsc{or} relations between the infinite planes that define each face of the polyhedron. The resulting disjunctive problem can be solved to global optimality using existing branch-and-bound techniques \cite{balas2018disjunctive}. This problem has a relatively high computational cost that grows exponentially with the number of obstacles \cite{balas2018disjunctive, ono2013probabilistic}. Even though recent efforts show promising improvements on computational efficiency, over-conservatism and probabilistic guarantees \cite{blackmore2011chance, ono2012closed, ono2013probabilistic, lefkopoulos2019using}, their computational cost is still too elevated to meet fast real-time requirements.

Alternatively, an obstacle can be bounded by a single differentiable surface (sphere, cylinder, ellipsoid, etc.) to be included as a nonlinear constraint of the optimization problem \cite{castillo2018model}. This results in a comparatively low-dimension nonlinear program (NLP), which can be solved efficiently by gradient-based solvers \cite{houska2011}. Even though this solution cannot guarantee global optimality, its reduced computational cost makes this strategy to be widely adopted in most time-critical motion planning tasks, such as model predictive control for aerial robots \cite{zhu2019chance,castillo2018model,kamel2017robust}. 

Accounting for uncertainty through a probabilistic framework has shown to overcome the inherent over-conservatism of set-bounded uncertainty models \cite{mesbah2016stochastic, ono2012closed}, which is essential to avoid unfeasibility in cluttered environments. However, the chosen strategy to bound the obstacles critically impacts the evaluation of the resulting chance constraints. For instance, the linear chance constraints that compose polyhedral obstacles have a closed-form deterministic equivalent for Gaussian systems \cite{blackmore2011chance}. On the other hand, nonlinear chance constraints need to be linearized \cite{zhu2019chance} or approximated by sampling methods \cite{blackmore2010probabilistic}, which leads to over-conservatism and high computational cost respectively. 

This paper proposes a hybrid solution that benefits from both strategies. First, a polyhedral obstacle formulation is exploited to provide a closed-form approximation of the disjunctive chance constraints. Then, a differential surface provides a safe bound on polyhedral obstacle regions. To meet fast real-time requirements, we restrict each polyhedral obstacle to be a cuboid (i.e. bounding box), and then obtain a tight quadratic bound analytically. As a result, we land on a nonlinear formulation that can be solved efficiently with the guarantee that the original chance constraints will be satisfied with the specified confidence level.


\section{Problem Statement}
\label{sec:problem_statement}



In this work, we consider the problem of motion planning with non-cooperative moving obstacles with uncertain localization, model and disturbances in the form of additive Gaussian noise. Thus, the dynamics of a given robot and a set of $N_o$ obstacles are described as the following stochastic, discrete-time model:
\begin{subequations}\label{eq:model}
  \begin{align}
    x_{t+1} &= f(x_t, u_t) + w_t \\
    y_{t+1}^i &= g^i(y_t^i) + v_t^i \qquad i \in \{1,\ldots, N_o\}
  \end{align}\label{eq:foo}
\end{subequations}
where $x_t \in \mathbb{R}^{n_x}$, $y_t^i \in \mathbb{R}^{n_y}$ and $u_t \in \mathbb{R}^{n_u}$ are the robot state, $i$-th obstacle state and robot inputs respectively at time $t \in \mathbb{N}$. $w_t \in \mathbb{R}^{n_w}$ and $v_t^i \in \mathbb{R}^{n_v}$ are unknown disturbances with Gaussian probability distributions; and $f$ and $g^i$ are (possibly nonlinear) Borel-measurable functions that describe the robot and the $i$-th obstacle dynamics respectively.

Let $p_t\subset x_t \in \mathbb{R}^3$ and $q_t^i \subset y^i_t \in \mathbb{R}^3$ be subspaces describing the position of their respective center of mass. Then, bounding boxes centered at $q_t^i$ with semi-sizes $d^i\in \mathbb{R}^3$ can be placed such that the free configuration space $\mathcal{F}_t$ is defined as follows:
\begin{equation}\label{eq:deterministic_collision}
  \mathcal{F}_t :=
  \left\lbrace
  x_t \in \mathbb{R}^{n_x} :
  \bigwedge_{i = 1}^{N_o}\bigvee_{j=1}^3 |p_t^{j} -q_t^{ij}|\geq d^{ij}
  \right\rbrace
\end{equation}
where $j$ iterates over the Cartesian coordinates of each $\mathbb{R}^3$ element. $|\cdot|$, $\bigvee$ and $\bigwedge$ denote the absolute value, the logical \textsc{or} and \textsc{and} respectively. Given the stochastic nature of the agents, we can define the chance constraint over the horizon length $N$ as follows:
\begin{equation}\label{eq:chance_constraint}
  \mathcal{P} \left( \bigwedge_{t = 1}^{N}x_t \in \mathcal{F}_t \right) \geq 1-\alpha
\end{equation}
which enforces the robot to stay within the free configuration space in a probabilistic sense with the confidence level $1-\alpha$. As a result, the probabilistic motion planning problem is defined as follows: 
\begin{subequations}\label{eq:ocp}
  \begin{flalign}
    \underset{u_0,\ldots, u_{N-1}}{\text{min.}} & J(u_0, \ldots, u_{N-1}, x_0, \ldots, x_N ) \label{eq:ocp_cost}\\
    \text{subject to:} &\nonumber \\
                       &x_{t+1} = f(x_t, u_t) + w_t \label{eq:robot_model_constraint}\\
                       &y_{t+1}^i = g^i(y_{t}^i) + v_{t}^i \label{eq:obstacle_model_constraint}\\
                       &w_t \sim \mathcal{N}(0, W_t) \quad
                       v_{t}^i \sim \mathcal{N}(0, V^i_t) \label{eq:robot_obstacle_uncertainty} \\
                       &x_0 \sim \mathcal{N}(\hat{x}_0, \Sigma_{x,0})\quad 
                        y_0 \sim \mathcal{N}(\hat{y}^i_0, \Sigma_{y,0})\label{eq:ocp_init}\\
                       &x_{t+1} \in \mathbb{X}, \quad u_t \in \mathbb{U} \label{eq:saturation}\\
                       &\mathcal{P} \left( \bigwedge_{t}x_{t+1} \in \mathcal{F}_{t+1} \right) \geq 1-\alpha \label{eq:ocp_obstacles}
   \end{flalign}
\end{subequations}
where $t \in \{0,\ldots, N-1\}$ and $i \in \{1, \ldots, N_o\}$. The cost function \cref{eq:ocp_cost} determines the objective to pursue such as energy consumption or a reference state. The stochastic model of the robot and the obstacles are included in equations \cref{eq:robot_model_constraint} to \cref{eq:robot_obstacle_uncertainty}. The initial states in \cref{eq:ocp_init} are assumed to be Gaussian distributions given by a state estimation algorithm such as Kalman filtering. The equations in \cref{eq:saturation} provide additional state and control constraints to be defined for a given application. Finally, the collision chance constraint is included in \cref{eq:ocp_obstacles} with confidence level $1-\alpha$.

The key difficulty of this problem lies on the evaluation of the non-convex chance constraint \cref{eq:ocp_obstacles}. It requires the integration of a multivariate Gaussian distribution and the convexification of the disjunctive constraints, which is, in general, intractable \cite{blackmore2011chance}. To overcome these difficulties, we safely approximate the problem as a deterministic disjunctive program, which is then casted into a nonlinear program to be solved efficiently by existing solvers \cite{houska2011}.

\section{Preliminary results}
\label{sec:existing}
For the sake of clarity, this section introduces preliminary results to support further developments in \cref{sec:diff_bound}. 
\subsection{Minimum volume enclosing ellipsoid of a bounding box}
Consider the space outside the bounding box $\mathcal{B}$ as 
\begin{equation}
  \mathcal{B}(d):=\left \lbrace x\in \mathbb{R}^3: \bigvee_{i=1}^3 |x_i| > d_i \right \rbrace 
\end{equation}
where $d \in \mathbb{R}_+^3$ and the Cartesian coordenates are iterated through the index $i$. This set can be safely approximated by its minimum volume enclosing ellipsoid $\mathcal{E}$, which can be computed in closed form as \cite{john2014extremum}:
\begin{equation}\label{eq:circumscribed}
  \mathcal{E}(d):=\left \lbrace x\in \mathbb{R}^3: \sum_{i=1}^3\left( \frac{x_i}{d_i}\right)^2 > 3 \right \rbrace
\end{equation}

\subsection{Chance constraints for linear-Gaussian systems}
Consider a multivariate Gaussian random variable $X \sim \mathcal{N}(\mu, \Sigma)$. Then, the chance constraint
\begin{equation}\label{eq:linear-gaussian}
  \mathcal{P}(a^TX + b \leq 0 ) \geq 1-\alpha, \qquad a,b \in \mathbb{R}^{n_x}
\end{equation}
has a deterministic equivalent of the form:
\begin{equation}\label{eq:linear-gaussian-result}
  a^T\mu + b + \Psi^{-1}(1-\alpha)\sqrt{a^T\Sigma a} \leq 0
\end{equation}
where $\Psi$ is the standard Gaussian cumulative distribution function defined as:
\begin{equation}
  \Psi(x) = \frac{1}{\sqrt{2\pi}} \int_{-\infty}^x exp \left \lbrace-\frac{t^2}{2}\right \rbrace
dt
\end{equation}
  

\subsection{Bounds on disjunctive chance constraints}
\label{sec:bounds}
As proven by \cite{ono2013probabilistic}, for any number of events $A_i$, we have:
\begin{equation}\label{eq:chance_union}
  \mathcal{P}
  \left( \bigvee_{i=1}^N A_i \right) \geq 1-\alpha 
  \Leftarrow 
  \bigvee_{i=1}^N \mathcal{P} \left( A_i \right) \geq 1-\alpha \\
\end{equation}
Similarly, as proven by \cite{ono2013probabilistic}, new variables $\alpha_i \in \mathbb{R}$ can be defined such that:
\begin{align}\label{eq:chance_intersection}
  &&
  \mathcal{P}
  \left( \bigwedge_{i=1}^N A_i \right) \geq 1-\alpha 
  \Leftarrow 
  \left( \bigwedge_{i=1}^N \mathcal{P} \left( A_i \right)
  \geq 1-\alpha_{i} 
  \right)\nonumber\\
  &&
  \wedge 
  \left( 0 \leq \alpha_i \leq 1 \right) 
  \wedge 
  \left(
    \sum_{i=1}^{N}\alpha_i \leq \alpha
  \right)
\end{align}
Thus, we have an immediate result on polyhedral obstacle regions described by chance constraints of the type: 
\begin{multline}\label{eq:chance_multiple}
  \mathcal{P}\left(
    \bigwedge_{t = 1}^N \bigwedge_{i=1}^{N_o}\bigvee_{j=1}^{N_f} A^{ij}_t 
\right) \geq 1-\alpha \Leftarrow\\
\bigwedge_{t = 1}^N \bigwedge_{i=1}^{N_o}\bigvee_{j=1}^{N_f}\mathcal{P}
\left( A^{ij}_t \right)\geq 1-\alpha^i_t\\
  \wedge 
  \left( 0 \leq \alpha^i_t \leq 1 \right) 
  \wedge 
  \left(
    \sum_{t=1}^{N}\sum_{i=1}^{N_o}\alpha^i_t \leq \alpha
  \right)
\end{multline}
where $N_f$ is the number of faces of the $i$-th obstacle. By direct comparison with recent results in \cite{jha2018safe,lefkopoulos2019using} we can see a considerable improvement on the chance constraint bounds, increasing the risk allocation parameters $\alpha^i_t$ by $N_f$ times for uniform risk allocation. Thus, less conservative bounds are obtained for the same confidence level, reducing the risk of posing unfeasible problems when multiple obstacles arise.

\section{Nonlinear bound for collision\\ chance constraints}
\label{sec:diff_bound}
This section develops the main theoretical contribution of this paper: a safe deterministic approximation of the chance constraint \cref{eq:chance_constraint} given by 
\small
\begin{multline}\label{eq:risk_ellipsoid}
  \bigwedge_{t = 1}^N \bigwedge_{i=1}^{N_o}\sum_{j=1}^3 \left(
    \frac{\hat{p}^{j}_t - \hat{q}^{ij}_t}{d^{ij}_t + \Psi^{-1}(1-\alpha^i_t)\sqrt{\sigma^2(p^{ij}_t) + \sigma^2(q^{ij}_t)}} 
  \right)^2
  \geq 3\\
  \wedge 
  \left( 0 \leq \alpha^i_t \leq 1 \right) 
  \wedge 
  \left(
    \sum_{t=1}^{N}\sum_{i=1}^{N_o}\alpha^i_t \leq \alpha
  \right)
\end{multline}
\normalsize
where $p^i_{t} \sim \mathcal{N}(\hat{p}^i_{t},\sigma^2(p^i_{t}))$ and $q^{ij}_{t} \sim \mathcal{N}(\hat{q}^{ij}_t,\sigma^2(q^{ij}_t))$. 

\begin{proof}
  Let the equation \cref{eq:chance_constraint} be rewritten as the disjunction:
\begin{equation}\label{eq:risk_constraint}
  \mathcal{P}\left(
  \bigwedge_{t = 1}^N \bigwedge_{i=1}^{N_o}\bigvee_{j=1}^3 \bigvee_{k=0}^1  
  (-1)^k(p^{j}_t -q^{ij}_t) + d^{ij}_t \leq 0
\right) \geq 1-\alpha
\end{equation}
By application of \cref{eq:chance_multiple}, we get:
\begin{align}\label{eq:chance_bound}
  &&
  \bigwedge_{t = 1}^N \bigwedge_{i=1}^{N_o}\bigvee_{j=1}^3 \bigvee_{k=0}^1  
  \mathcal{P}\left(
  (-1)^k(p^{j}_t -q^{ij}_t) + d^{ij}_t \leq 0
\right) \geq 1-\alpha^{i}_t \nonumber\\
  &&
  \wedge 
  \left( 0 \leq \alpha^i_t \leq 1 \right) 
  \wedge 
  \left(
    \sum_{t=1}^{N}\sum_{i=1}^{N_o}\alpha^i_t \leq \alpha
  \right)
\end{align}

Since we now have linear combinations of Gaussian variables we can apply equation \cref{eq:linear-gaussian-result} to obtain: 
\begin{multline}
  \Bigg(
  \bigwedge_{t = 1}^N \bigwedge_{i=1}^{N_o}\bigvee_{j=1}^3 \bigvee_{k=0}^1  
(-1)^j(\hat{p}^j_t - \hat{q}^{ij}_t) + d^{ij}_t\\ 
+ \Psi^{-1}(1-\alpha_t^i)\sqrt{\sigma^2(p^{j}_t) + \sigma^2(q^{ij}_t)} < 0 \Bigg)\\
\wedge 
\left( 0 \leq \alpha^i_t \leq 1 \right) 
\wedge 
\left(
  \sum_{t=1}^{N}\sum_{i=1}^{N_o}\alpha^i_t \leq \alpha
\right)
\end{multline}
which is equivalent to 
\begin{multline}\label{eq:risk_box}
  \Bigg( 
  \bigwedge_{t = 1}^N \bigwedge_{i=1}^{N_o}\bigvee_{j=1}^3 
  |\hat{p}^j_t -\hat{q}^{ij}_t|\geq d^{ij}_t \\ 
  + \Psi^{-1}(1-\alpha_t^i)\sqrt{\sigma^2(p^j_t) + \sigma^2(q^{ij}_t)} \Bigg)\\
  \wedge 
  \left( 0 \leq \alpha^i_t \leq 1 \right) 
  \wedge 
  \left(
    \sum_{t=1}^{N}\sum_{i=1}^{N_o}\alpha^i_t \leq \alpha
  \right)
\end{multline}
The equation \cref{eq:risk_box} defines a bounding box to which \cref{eq:circumscribed} can be applied to obtain \cref{eq:risk_ellipsoid} and complete the proof.
\end{proof} \vspace{-0.2cm}
This methodology allows the problem \cref{eq:ocp} to be addressed through nonlinear programming, which critically impacts its tractability and the scalability. For instance, each polyhedral obstacle requires $7N$ mixed-integer constraints and $6N$ binary variables \cite{lefkopoulos2019using}, while our method can be implemented with $N$ quadratic constraints and zero additional variables. In addition, the disjunctive program has a relatively high computational cost that grows exponentially with the number of obstacles \cite{ono2013probabilistic,balas2018disjunctive}. In contrast, our nonlinear program can be solved with polynomial complexity \cite{houska2011}, being computationally efficient for large-scale problems \cite{biegler2009large}. 

\section{Case Study: Robot Collision Avoidance}
\label{sec:case_study}
In this section we implement our motion planning approach \cref{eq:ocp} in a Model Predictive Control (MPC) fashion to provide collision-free navigation on a DJI-M100\footnote{DJI Matrice 100: \url{https://www.dji.com/matrice100}} quadrotor. The results of the experiments are complemented by the video demonstration \url{https://rebrand.ly/castillo_RAL2020}. 

\subsection{Robot Model}
Based on the DJI SDK, the control inputs given to the quadrotor are defined as $u = [ u_x\ u_y\ u_z\ u_\psi]^T$, which correspond to forward, sideward, upward, and heading velocity references, respectively based on a local frame $L$ parallel to the ground (see \cite{castillo2018model} for details). Thus, the nominal system dynamics are modeled as follows:
\begin{subequations}\label{eq:con_model}
  \begin{align}
    \dot{p} &= R(\psi)v\\
    \dot{v}_i &= \frac{1}{\tau_i}(-v_i + k_i u_i) \label{eq:fom}, \quad i \in \lbrace x,y,z \rbrace\\
    \ddot{\psi} &= \frac{1}{\tau_{\psi}}(-\dot{\psi} + k_{\psi} u_\psi) \label{eq:fom}
  \end{align}\label{eq:foo}
\end{subequations}
where $v=[ v_x\ v_y\ v_z]^T$ is the linear velocity of the center of mass in the local frame and $R(\psi)$ the rotation matrix for the yaw angle $\psi$. $k_i$, $k_{\psi}$ and $\tau_i$, $\tau_{\psi}$ are the gain and time constants relative to each component of $u$ respectively. Thus, the robot state is defined as $x_t = [p_t\ v_t\ \psi_t\ \dot{\psi}_t]$ and the nominal discrete dynamics $f(x_t,u_t)$ are obtained through 4-th order Runge-Kutta integration of \cref{eq:con_model}. The nominal state prediction $\hat{x}_t$ and its covariance matrix $\Sigma^x_t$ are approximated with a first-order Taylor expansion \cite{luo2017review}: 
\begin{subequations}\label{eq:robot_model}
  \begin{align}
    &\hat{x}_{t+1} = f(\hat{x}_t, u_t)\\
    &\Sigma^{x}_{t+1} = 
    \left( \nabla_x f(\hat{x}_t, u_t)\right)
    \Sigma^{x}_{t} 
    \left( \nabla_x f(\hat{x}_t, u_t)\right)^T 
    + W_t
  \end{align}
\end{subequations}
where $u_t$ is obtained from the predicted inputs of the MPC algorithm. Even though there exists more precise uncertainty propagation methods \cite{luo2017review}, we use Taylor expansion for the sake of computational efficiency.

\subsection{Obstacle Model}
Obstacles are modeled with constant velocity nominal dynamics:
  \begin{equation}\label{eq:con_obs_model}
    \dot{q}^i = R(\psi^i)v^i,\qquad \dot{v}^i = \ddot{\psi}^i = 0
  \end{equation}
  where $v^i$ and $\psi^i$ are the linear velocity in the body frame and yaw angle of the $i$-th obstacle respectively. Thus, obstacle states are defined as $y^i_t =[q^i_t\ v^i_t\ \psi^i_t\  \dot{\psi}^i_t]$ where the nominal discrete dynamics $g^i(y^i_t)$ are determined through Euler integration of \cref{eq:con_obs_model}. Similarly, the nominal state $\hat{y}^i_t$ and its covariance matrix $\Sigma^{y^i}_{t}$ are approximated with a first-order Taylor expansion
  \begin{subequations}\label{eq:obstacle_model}
  \begin{align}
    &\hat{y}^i_{t+1} = g^i(\hat{y}^i_t)\\
    &\Sigma^{y^i}_{t+1} = 
      \nabla g^i(\hat{y}^i_t)
      \Sigma^{y^i}_{t}
      \left(\nabla g^i(\hat{y}^i_t)\right)^T
    + V^i_t
  \end{align}
\end{subequations}

\subsection{Objective Function}
We define the cost function in \cref{eq:ocp_cost} as:
\begin{equation}\label{eq:particular_cost}
  J = \sum_{t=1}^N 
  \left( 
    \| x_t - x^r_t\|^2_P +
    \| u_{t-1}\|^2_Q
  \right)
\end{equation}
where $x^r_t$ is the user-defined goal state. $\|\cdot\|_P$ and $\|\cdot \|_Q$ are the norms induced by the $P$ and $Q$ weighting matrices.

\subsection{One-Horizon Benchmark}
\begin{figure}[hbtp]
    \centering
    \includegraphics[width=\columnwidth]{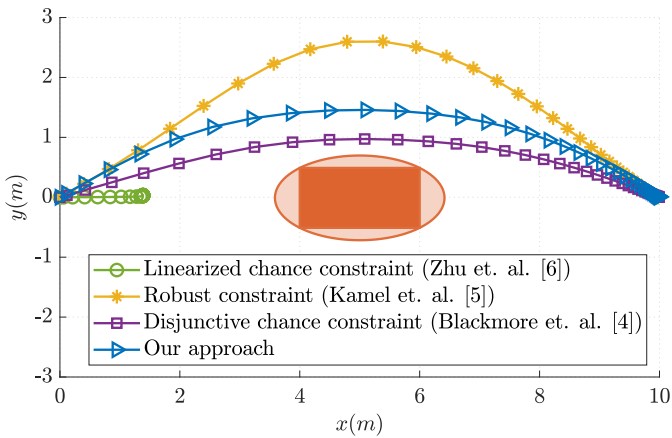}
    \caption{One horizon benchmark of our approach against the linearized chance constraint from \cite{zhu2019chance}, the robust constraint from \cite{kamel2017robust} and the disjunctive chance constraint from \cite{blackmore2011chance}.}
    \label{fig:benchmark}
  \end{figure}

In this section, our method is compared against three state-of-the-art approaches \cite{kamel2017robust, zhu2019chance, blackmore2011chance} on stochastic optimal collision avoidance for real-time systems. We design a two-dimensional experiment where the robot and the obstacle are placed at $p_0=[0\ 0]$ and $q_0 = [5\ -0.01]$ respectively. Uncertain obstacle's location is considered with covariance $\Sigma_q = diag(0.4\ 0.1)$. The bounding box size is $d = [1\ 0.5]$ as shown in \cref{fig:benchmark}. We have selected a prediction horizon of $8$ seconds with $N = 40$ steps and a confidence level $ 1-\alpha = 0.99$ with uniform risk allocation $\alpha^i_t = \alpha /N$. For the sake of a purely chance-constrained benchmark, we have dropped the additional potential fields implemented in \cite{kamel2017robust, zhu2019chance} that would have made these implementations even more conservative.

As shown in \cref{fig:benchmark}, our approach avoids the tendency of linearized chance constraints \cite{zhu2019chance} to fall into local minima while providing a level of conservatism between the robust constraint from \cite{kamel2017robust} and the disjunctive chance constraint from \cite{blackmore2011chance}. As presented in \cref{table:benchmark}, our approach solves a conservative approximation of \cite{blackmore2011chance} over 142 times faster at the price 4\% of optimality. Our computation time falls in the range of \cite{kamel2017robust,zhu2019chance}, which have been widely used for fast real-time motion planning. The one-horizon benchmark has been executed from the optimization framework CasADi \cite{Andersson2018}, being publicly available on-line to be reproduced\footnote{ Benchmark code: \url{https://rebrand.ly/castillo_RAL2020benchmark}}.

\begin{table}[hbtp]
    \caption{Relative results from the one horizon benchmark.}
		\label{table:benchmark}
		\centering
    \begin{tabular}{l|cccc}
      & Ours & Kamel\cite{kamel2017robust} & Zhu\cite{zhu2019chance} & Blackmore\cite{blackmore2011chance}\\
      \hline \\ [-1.9ex]
      Objective     & 1.0 & 1.0925  & 2.6166               & 0.9614 \\
      CPU time (s)  & 1.0 & 1.3198  & 0.4555 \footnotemark & 142.08\\
		\end{tabular}
     \vspace{-1em}
\end{table}
\footnotetext{The solver converges faster when falling into local minima.}

\subsection{Real experiment: Pedestrian collision avoidance.}\label{sec:real}
The experiment consists in two pedestrians who naturally walk inside a closed area where the robot is operating. As shown in \cref{fig:experiment_snapshot} and the complementary video, when the pedestrians intend to occupy the robot's safe space, evasive trajectories are planned and executed while tracking a reference position given by $p^r_t= [0\ 0\ 1.5]\ m$.

The experiment is conducted in a flying arena of $[4\ 3\ 3]\ m$ equipped with an Optitrack\footnote{Optitrack motion capture system \url{https://optitrack.com/}} motion capture system, which provides raw pose measurements of the robot and the obstacles. These poses are processed by Extended Kalman Filter (EKF) algorithms \cite{sanchez2017visual} according to the robot \cref{eq:robot_model} and the obstacle \cref{eq:obstacle_model} models. Gaussian model disturbances in linear and angular velocities have been considered as $\sigma^2(v_t) = 0.03\ m^2/s^2$ and $\sigma^2(\dot{\psi}_t) = 0.03\ rad^2/s^2$ for the robot and the obstacles. The measurement noise on position has been identified to be $\sigma^2(p_t) =\sigma^2(q^i_t) = 2.5\cdot 10^{-3} m^2$. The bounding boxes around the pedestrians are defined by $d^i_t = [2\ 2\ 4]\ m$ with confidence level $ 1-\alpha = 0.99$ and uniform risk allocation $\alpha^i_t = \alpha /NN_o$. The real-time implementation of the problem \cref{eq:ocp} with $N=20$ steps over $4s$ of prediction horizon is based on ACADO Toolkit \cite{houska2011} and ROS Kinetic\cite{quigley2009ros} C++ framework running on a on an Intel i7-6820HQ CPU@2.70GHz.

\begin{figure}[hbtp]
  \vspace{-0.2cm}
  \centering
    \includegraphics[width=\columnwidth]{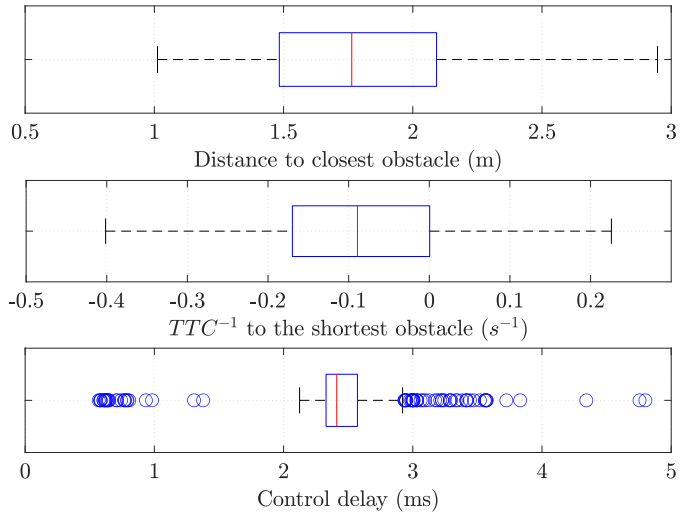}
    \caption{Pedestrian Collision Avoidance: Box plots for the distance to the closest obstacle, the inverse time to collision $TTC^{-1}$, and the control delay. The median is represented in red and the 25-75th percentiles in blue. The black whiskers represent 1.5 times the interquartile range. Outliers are plotted as blue circles}
    \label{fig:box_plots}
  \vspace{-0.5cm}
\end{figure}

In this work, we include the results over 5 minutes of experiment. The outcome of this experiment in terms of safety are evaluated statistically through the distance to the closest obstacle $d$ and its inverse time-to-collision $TTC^{-1} = \dot{d}/d$ \cite{van1993time}. Large negative values of $TTC^{-1}$ indicate high risk of collision, while values near zero correspond to safe situations \cite{van1993time}. As shown in \cref{fig:box_plots}, the robot presented a low risk of collision, since the distance to the closest obstacle lies in the range $[1,3]\ m$ with median $1.7\ m$ and the $TTC^{-1}$ values are concentrated around  $-0.09s^{-1}$ with a minimum value of $-0.4\ s^{-1}$. In addition, our approach presents fast real-time capabilities with a median control delay of $2.4\ ms$.
\subsection{Simulation: Crowd Collision Avoidance}
\begin{figure}[hbtp]
  \centering
  \includegraphics[width=\columnwidth]{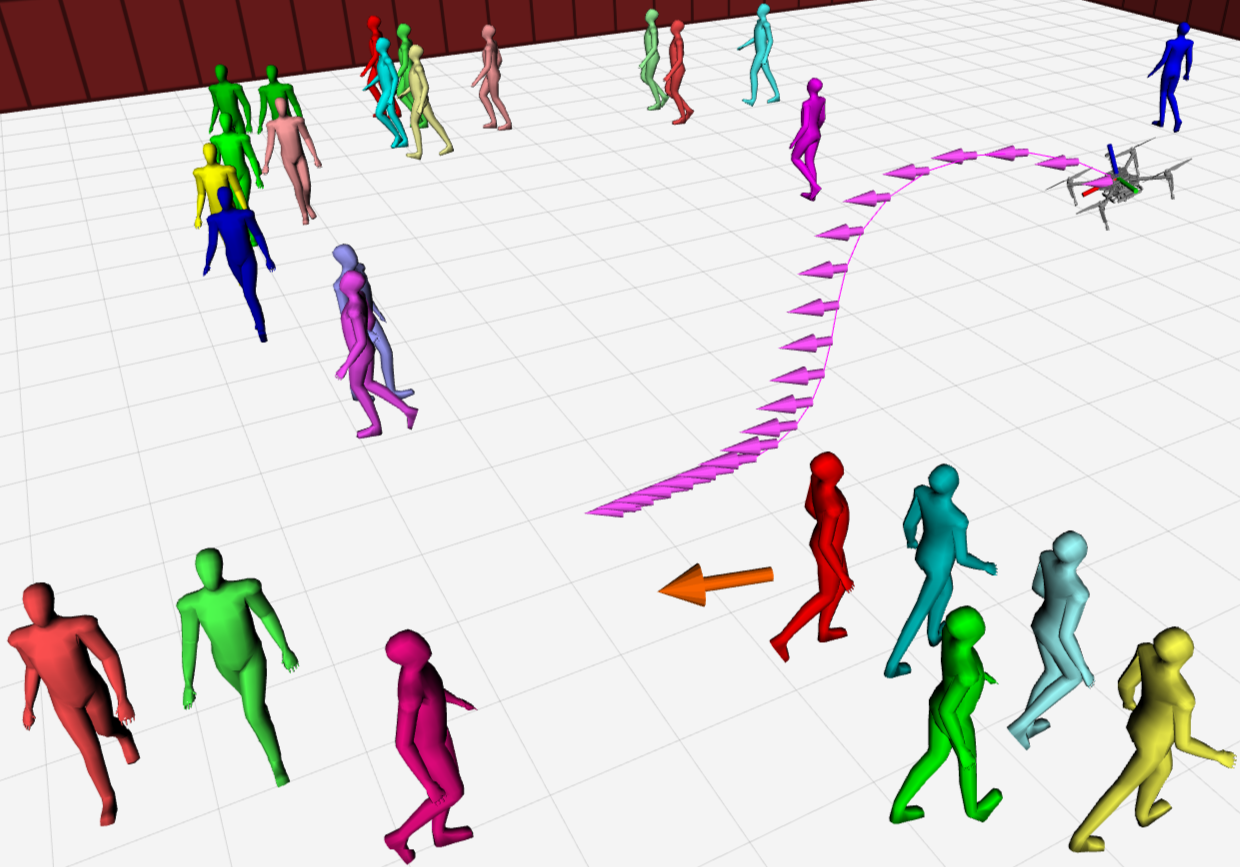}
    \caption{Crowd collision avoidance simulation with 30 pedestrians. The orange arrow represents the moving reference position. The robot pose and predicted trajectory are indicated by the frame and the purple arrows respectively.}
    \label{fig:pedsim_rviz_screenshot}
\end{figure}

This experiment consists in a software-in-the-loop simulation where the robot navigates in a crowded scenario. 30 pedestrians, driven by the social force model \cite{helbing1995social}\footnote{Pedestrian simulator code: \url{https://github.com/srl-freiburg/pedsim_ros}}, follow a squared path of $14\ m$ length with a reference velocity of $1\ m/s$. The robot, simulated according to \cref{eq:con_model}, is tracking the same path in opposite direction at $1.5 m/s$ while avoiding the pedestrians, as shown in \cref{fig:pedsim_rviz_screenshot} and the complementary video. The simulation runs at $100\ Hz$ with the same setup as the experiment conducted in \cref{sec:real}.

\begin{figure}[hbtp]
  \centering
  \includegraphics[width=\columnwidth]{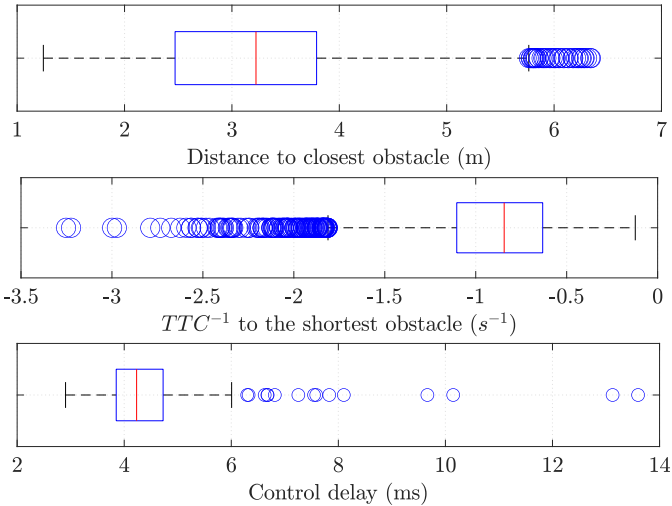}
    \caption{Crowd Collision Avoidance: Box plots for the distance to the closest obstacle, the inverse time to collision $TTC^{-1}$, and the control delay. The median is represented in red and the 25-75th percentiles in blue. The black whiskers represent 1.5 times the interquartile range. Outliers are plotted as blue circles}
    \label{fig:pedsim_box_plots}
    \vspace{-0.5cm}
\end{figure}

In this simulation we include the results over 20 minutes of experiment. Analogously to \cref{sec:real}, the outcome of this experiment is evaluated statistically through the inverse time to collision ($TTC^{-1}$), the distance to the closest obstacle and the control delay, as shown in \cref{fig:pedsim_box_plots}. The nature of the experiment and the higher number of obstacles involves a greater risk than the previous experiment, with a median $TTC^{-1}$ of $-0.84\ s^{-1}$. Consequently, our algorithm shows a more conservative behavior, with a median distance to the closest obstacle of $3.22\ m$. Finally, the higher number of obstacles moderately increases the computation time to $4.2\ ms$, leaving room to scale up to more complex scenarios.


\section{Conclusions} \label{sec:conclusions}

We presented a new real-time approach to address chance-constrained motion planning with dynamic obstacles. The obstacles are considered to have uncertain localization, model and disturbances in the form of additive Gaussian noise. We developed a closed-form differentiable bound on the probability of collision to safely approximate the disjunctive chance-constrained optimization problem as a nonlinear program. Consequently, the computational cost was reduced dramatically while maintaining the original safety guarantees, allowing its implementation in fast real-time applications. Through mathematical proof and simulations, our method has shown to reduce conservatism with respect to recent real-time approaches, remaining tractable when accounting for multiple obstacles. Finally, real-time experiments validated the presented approach using nonlinear model predictive control on an aerial robot to avoid pedestrians. Future work will consider closed-loop constraint satisfaction techniques \cite{ono2012closed} and alternative risk-allocation methods \cite{ono2013probabilistic} to further reduce conservativeness while maintaining the required safety guarantees. In addition, new practical applications will be targeted, including other robotic platforms, perception and energy consumption objectives.

\bibliography{./main}
\bibliographystyle{IEEEtran.bst}

\end{document}